\documentclass[sigconf,nonacm]{acmart}
\usepackage{algorithm}
\usepackage{algpseudocode}
\usepackage{float}

\AtBeginDocument{%
  \providecommand\BibTeX{{%
    \normalfont B\kern-0.5em{\scshape i\kern-0.25em b}\kern-0.8em\TeX}}}

\setcopyright{acmlicensed}
\copyrightyear{2026}
\acmYear{2026}
\acmDOI{XXXXXXX.XXXXXXX}

\graphicspath{{./images/}} 

\begin{document}

\title{Adaptive Ensemble Selection for Noisy Labels on Tabular Data}

\author{Faizaan Ali}
\orcid{0009-0001-2187-6728}
\affiliation{
  \institution{Rensselaer Polytechnic Institute}
  \city{Troy}
  \state{New York}
  \country{USA}
}
\email{alif2@rpi.edu}

\author{Inwon Kang}
\affiliation{%
	\institution{Rensselaer Polytechnic Institute}
	\city{Troy}
	\state{New York}
	\country{USA}}
\email{kangi@rpi.edu}

\author{Oshani Seneviratne}
\affiliation{%
	\institution{Rensselaer Polytechnic Institute}
	\city{Troy}
	\state{New York}
	\country{USA}}
\email{senevo@rpi.edu}

\renewcommand{\shortauthors}{Ali, Kang and Seneviratne}

\begin{abstract}
Incorrect or corrupted labels in tabular datasets can significantly degrade supervised learning performance, particularly when mislabeling is subtle and not easily detectable from feature space alone. In the context of automated or AI-augmented data science workflows, robust detection of such label noise is critical for building reliable models. 
We propose a data-centric reasoning module for AI data science systems that automatically diagnoses dataset quality and selects appropriate cleaning strategies.
Given a dataset, a meta-model predicts weights over a diverse set of detectors, including confidence-based, neighborhood-based, and distributional methods. Across benchmark datasets with controlled noise, our approach achieves performance comparable to a Confident Learning baseline on average, with dataset-dependent gains and losses, particularly in heterogeneous regimes. We further show that detector effectiveness is systematically linked to dataset properties. These results demonstrate the value of descriptor-driven, data-centric ensembling as a component of AI-assisted data-science pipelines for robust dataset assessment and model reliability.
\end{abstract}



\maketitle

\section{Introduction}

Label noise is a fundamental challenge in supervised learning, arising when observed target labels do not reflect the true underlying class of an instance. Such noise can result from annotation errors, ambiguous labeling guidelines, or class overlap. In automated or AI-augmented data science workflows, mislabeled examples distorts model objectives, leading to incorrect pattern learning and degraded generalization. This challenge is especially pronounced in tabular data, where heterogeneous features lack the spatial or semantic structure present in vision or language tasks. Mislabeled instances often occur near class boundaries or in sparsely populated regions, making them difficult to distinguish from genuinely hard examples \cite{nazaretyan2025benchmarking}.

A variety of label noise detection methods exist, including confidence-based filtering \cite{nazaretyan2025benchmarking}, nearest-neighbor consistency \cite{garcia2012study}, and probabilistic approaches, such as Confident Learning \cite{northcutt2021confident}. However, prior work \cite{nazaretyan2025benchmarking, garcia2012study} demonstrates that no single method is universally optimal. Detector performance depends strongly on dataset characteristics, such as class imbalance, feature dimensionality, and distributional properties.

In this work, we adopt a data-centric, meta-learning perspective aligned with AI data science systems: can label noise detection be dynamically adapted to the dataset itself? We propose an adaptive ensemble framework where a meta-model maps dataset descriptors to weights over a diverse set of detectors, enabling dynamic aggregation tailored to the underlying data regime. This approach can be viewed as an automated, intelligent module that augments human data scientists  or AI data science systems in evaluating dataset quality and improving model reliability.
This meta-model can be integrated with LLM-based agents that extract dataset descriptors or reason about cleaning strategies.

We evaluate the approach on benchmark tabular datasets with controlled noise and demonstrate performance that is competitive with the baseline, with dataset-dependent gains and losses. Further analysis reveals that detector performance correlates systematically with dataset properties, and that the meta-model learns interpretable weighting strategies aligned with these relationships.

Our contributions are:
\begin{itemize}
\item We formulate label noise detection as a meta-learning problem over dataset descriptors, enabling adaptive, dataset-aware decision-making.
\item We propose an adaptive ensemble that dynamically weights complementary detectors, suitable for integration into AI-assisted data-science pipelines.
\item We show that descriptor-driven ensembling can improve robustness in some dataset and noise regimes.
\item We provide empirical and interpretable evidence linking dataset properties to detector effectiveness, offering insights for intelligent system design in data science workflows.
\end{itemize}

\section{Related Work}

The problem of identifying mislabeled data has been widely studied in the context of improving dataset quality and downstream model performance \cite{garcia2012study, northcutt2021confident, srikanth2023empiricalstudyautomatedmislabel}. 
Nazaretyan et al.~\cite{nazaretyan2025benchmarking} systematically benchmark label-error detection methods for tabular data, including classifier-, neighborhood-, and ensemble-based approaches.
Their results show that no single method consistently dominates across datasets and noise conditions, and that mislabeled instances often lie in the fringe near class boundaries, motivating the use of local inconsistency and confidence-based signals.
Early work by Rebbapragada et al.
~\cite{rebbapragada2012active} introduced active label correction, in which likely errors are iteratively selected for expert review.

More recent work by Northcutt et al.~\cite{northcutt2021confident} introduces the Confident Learning framework, which estimates the joint distribution between noisy and latent true labels using out-of-sample predicted probabilities. Confident Learning enables principled identification and ranking of mislabeled instances and has become a widely adopted baseline through its Cleanlab implementation due to its strong empirical performance and ease of integration.

Despite these advances, both benchmarking studies~\cite{nazaretyan2025benchmarking} and probabilistic approaches~\cite{northcutt2021confident} highlight a key limitation: detector effectiveness is highly dependent on dataset characteristics, such as class imbalance, feature structure, and noise type. Supporting this observation, Srikanth et al.~\cite{srikanth2023empiricalstudyautomatedmislabel} show that even strong pipelines built on Confident Learning vary significantly across datasets and noise regimes. Furthermore, Garcia et al.~\cite{garcia2012study} demonstrate that heterogeneous ensemble-based filtering strategies based on majority consensus can improve robustness but exhibit inconsistent gains across settings.

These findings underscore a central challenge: no single detector or fixed aggregation strategy performs reliably across heterogeneous datasets. In contrast, our work models detector selection as a function of dataset-level characteristics, learning to adaptively weight detectors based on meta-features to better capture this variability.

The present work also builds on broader data-centric and task-adaptive approaches to tabular learning. Seneviratne et al. frame data excellence as a first-class requirement for data-centric AI~\cite{seneviratne2021towards} and examine explanation-driven quality assessment in rule-based systems~\cite{seneviratne2025explainability}. Kang et al. study representation learning and benchmarking for tabular data distillation, emphasizing the heterogeneity of tabular features and downstream learners~\cite{kang2025on}. More recently, they use handcrafted task descriptors in a meta-learner that adapts softmax calibration across few-shot tabular classification tasks~\cite{kang2026language}. Our setting differs in its objective: rather than adapting a classifier or a distilled representation, we learn dataset-conditioned weights over label-error detectors. These works nevertheless motivate the use of dataset descriptors to adapt data-processing and learning strategies across heterogeneous tabular regimes.

\section{Methodology}

\subsection{Problem Formulation}
Let $\mathcal{D} = \{(\mathbf{x}_i, y_i)\}_{i=1}^{N}$ be a classification dataset where each label $y_i \in \{1, \ldots, C\}$ may have been corrupted by an unknown noise process. We define a label noise detector as a function $f: (\mathbf{X}, \mathbf{y}) \rightarrow \mathbf{s} \in [0,1]^N$ that assigns a scalar suspicion score to every sample, where higher values indicate a greater likelihood of mislabeling. These scores may represent probabilities or normalized anomaly measures depending on the detector. Given a portfolio of $K$ detectors $\mathcal{F} = \{f_1, \ldots, f_K\}$, our goal is to learn a meta-function $g: \phi(\mathcal{D}) \rightarrow \mathbf{w} \in \mathbb{R}^K$ that maps dataset-level meta-features $\phi(\mathcal{D})$ to a non-negative weight vector satisfying $\sum_{k=1}^K w_k = 1$. The resulting ensemble score is given by $\mathbf{s}^* = \sum_{k=1}^K w_k f_k(\mathbf{X}, \mathbf{y})$, which aims to maximize detection performance across diverse dataset regimes.

\subsection{Fringe-Based Noise Injection}
To generate controlled ground-truth noise for meta-model training, we adopt a \textit{fringe-based} injection strategy as shown in Algorithm \ref{alg:fringe} rather than using uniform random label flipping. The motivation is that real-world annotation errors are disproportionately concentrated near decision boundaries, where examples are ambiguous and annotators are least confident~\cite{nazaretyan2025benchmarking}.

For a given noise rate $\varepsilon \in (0,1)$, we first fit a logistic regression model on the standardized feature matrix $\mathbf{X}$ and extract the predicted probability $p_i = \hat{P}(y_i \mid \mathbf{x}_i)$ for each sample under its current label. Let $N = |\mathcal{D}|$ denote the number of samples. The $\lfloor N \cdot \varepsilon \rfloor$ samples with the lowest confidence scores (i.e., assigned the lowest observed-label confidence) form the fringe set $\mathcal{F}_\varepsilon$. Each sample $i \in \mathcal{F}_\varepsilon$ has its label replaced by a class drawn uniformly at random from the remaining classes. The set of flipped indices is retained as the ground-truth noise mask for evaluation. We apply this procedure uniformly across both meta-model training and benchmarking at noise rates $\varepsilon \in \{0.05, 0.10, 0.20, 0.30, 0.40, 0.50\}$. The noise rate $\varepsilon$ specifies the fraction of samples that are selected for label flipping from the lowest-confidence fringe. It does not imply that every sample has an independent probability $\varepsilon$ of being corrupted. Consequently, the corruption sets are nested: as \(\epsilon\) increases, they expand from the lowest-confidence region.

\begin{algorithm}[t]
\caption{Fringe-Based Label Noise Injection}
\label{alg:fringe}
\begin{algorithmic}[1]
\Require Feature matrix $\mathbf{X} \in \mathbb{R}^{N \times d}$, labels $\mathbf{y} \in \mathcal{C}^N$, noise rate $\varepsilon$
\Ensure Noisy labels $\mathbf{y'}$, flipped index set $\mathcal{F}_\varepsilon$

\State $N \leftarrow |\mathbf{y}|$
\State Standardize features: $\mathbf{X_s} \leftarrow \text{StandardScaler}(\mathbf{X})$
\State Encode labels to integers
\State Train logistic regression classifier on $(\mathbf{X_s}, \mathbf{y})$
\State Compute class probabilities $P(y \mid \mathbf{x}_i)$ for each sample
\State Compute self-confidence scores $p_i = P(y_i \mid \mathbf{x}_i)$
\State $n \leftarrow \lfloor N \cdot \varepsilon \rfloor$
\State $\mathcal{F}_\varepsilon \leftarrow$ indices of the $n$ smallest $p_i$
\State $\mathbf{y'} \leftarrow \mathbf{y}$

\For{each $i \in \mathcal{F}_\varepsilon$}
    \State Sample $y'_i$ uniformly from $\{c \in \mathcal{C} \mid c \neq y_i\}$
\EndFor

\State \Return $\mathbf{y'}$, $\mathcal{F}_\varepsilon$
\end{algorithmic}
\end{algorithm}

\subsection{Label Noise Detector Suite}
To identify mislabeled samples, we assemble a diverse suite of eight complementary detectors, each reflecting a distinct inductive bias about what constitutes a suspicious label. Our goal is for the ensemble to capture multiple facets of label noise: some detectors exploit model uncertainty (low-confidence predictions), others measure local or global consistency (neighbor or graph agreement), and others detect distributional anomalies that may indicate outliers. By combining different approaches, the suite balances sensitivity to subtle mislabeling with robustness against spurious signals. Next, we describe each detector in detail, highlighting their assumptions, mechanisms, and key hyperparameters.

\subsubsection{Out-of-Fold Confidence Detectors}
Three supervised classifiers, logistic regression (\texttt{logreg\_conf}), random forest (\texttt{rf\_oof}), and histogram-based gradient boosting (\texttt{gb\_conf}), are each trained under stratified $K$-fold cross-validation ($K=3$). For each held-out fold, predicted class probabilities are recorded, and the per-sample suspicion score is defined as $s_i = 1 - \hat{P}(y_i \mid \mathbf{x}_i)$ capturing the idea that a consistently low posterior under the observed label is evidence of mislabeling. To mitigate within-fold overfitting, we apply $\ell_2$ regularization ($C{=}0.5$) to the logistic regression, enforce \texttt{min\_samples\_leaf}{$=$}3 in the random forest, and use $\ell_2$ regularization with early stopping in the gradient boosting model.

\subsubsection{K-Nearest Neighbor Inconsistency}
The KNN detector trains a $K$-nearest neighbor classifier ($K{=}15$) on the full dataset and flags samples where the predicted label disagrees with the observed label. This yields a binary suspicion score that captures local neighborhood inconsistency without requiring a held-out partition.

\subsubsection{Density-Based Anomaly Detectors}
Two unsupervised, label-agnostic detectors capture distributional outliers. Isolation Forest~\cite{liu2008isolation} partitions the feature space via random axis-aligned splits and assigns anomaly scores based on the mean path length required to isolate a sample. Local Outlier Factor~\cite{breunig2000lof} computes the ratio of a sample's local reachability density to that of its $K$-nearest neighbors ($K{=}20$), flagging samples that reside in regions significantly less dense than their surroundings. Both detectors return binary flags ($1$ for anomaly, $0$ otherwise).

\subsubsection{Graph-Based Label Propagation}
The Label Propagation detector exploits semi-supervised learning principles to identify labels that are inconsistent with the graph structure of the feature space. A $k$-nearest neighbor graph ($k{=}15$) is constructed over the standardized features, and a Label Spreading model~\cite{zhou2003learning} propagates label information across graph edges with a clamping parameter $\alpha{=}0.2$, allowing labels to diffuse smoothly through connected regions. For each sample, the detector computes the discrepancy between the observed label and the graph-smoothed posterior distribution:
\begin{equation}
    s_i = 1 - \tilde{P}(y_i \mid \mathbf{x}_i, \mathcal{G}),
\end{equation}
where $\tilde{P}$ denotes the propagated label distribution and $\mathcal{G}$ is the feature-space graph. This approach is particularly effective when mislabeled samples reside in neighborhoods dominated by a different class, as the graph consensus overrides the corrupted observation.

\subsubsection{Confident Learning as a Detector}
We additionally include Confident Learning~\cite{northcutt2021confident} as an individual detector within the ensemble. Using out-of-sample predicted probabilities from stratified cross-validation, Confident Learning produces per-sample label quality estimates $q_i \in [0,1]$ based on normalized prediction margins. 

To align with the rest of our framework, we convert these into suspicion scores via
\begin{equation}
    s_i = 1 - q_i.
\end{equation}

This allows Confident Learning to function both as a strong standalone baseline (Section~3.6) and as a complementary signal within the ensemble.

\subsection{Dataset Meta-Features}
To capture relevant characteristics of each dataset, we summarize it using a 19-dimensional descriptor vector $\phi(\mathcal{D}) \in \mathbb{R}^{19}$, grouped into four categories. These meta-features are designed to provide the label noise detectors with information about class structure, feature geometry, distributional properties, and signal strength, which can all influence the detectability of mislabeled samples.

\textbf{Class distribution} (4 features): The number of classes $C$, the proportion of the majority class, the Shannon entropy of the class labels, and the imbalance ratio (maximum-to-minimum class count). These features describe the overall label balance, which can affect how easily a mislabeled point stands out: for example, noise in a highly imbalanced dataset may be harder to detect without considering the class proportions.

\textbf{Dataset geometry} (3 features): The number of features $d$, the log-transformed sample count $\log(1+N)$, and the sample-to-feature ratio $N/d$. These geometric properties inform the model about dataset scale and dimensionality, helping to contextualize distances, sparsity, and density estimates used in neighbor- and graph-based detectors.

\textbf{Feature statistics} (7 features): The mean feature variance, fraction of near-constant features (variance ${<}10^{-5}$), sparsity (fraction of zero entries), mean absolute skewness, and pairwise correlation statistics (mean, maximum, and fraction of correlations $|\rho| > 0.8$). These statistics highlight the structure and redundancy in the feature space: for example, highly correlated or near-constant features may reduce effective dimensionality, affecting detectors that rely on distances or density.

\textbf{Signal structure} (5 features):
The between-class/within-class variance ratio; the explained-variance ratios of the first three principal components; and their cumulative explained variance.

All features are sanitized by replacing \texttt{NaN} and infinite values with zero prior to model input. By providing this mixture of descriptive statistics and structure-aware metrics, the meta-features give the label noise detectors a richer understanding of the dataset, improving their ability to identify subtle mislabeling patterns.

\subsection{Meta-Model Architecture and Training Objective}

Given a dataset descriptor $\phi(\mathcal{D})$, the meta-model predicts a set of weights over the base detectors. These weights reflect the relative reliability of each detector for the dataset at hand, enabling the ensemble to adapt to different data regimes. In essence, the meta-model learns to reason about which detectors are likely to be most effective, based on dataset-level characteristics, without requiring access to the ground-truth labels at inference time.

\subsubsection{Architecture}
The meta-model $g_\theta$ is a two-layer feed-forward network. Let $\mathbf{x} \in \mathbb{R}^{d_\phi}$ denote the input descriptor vector (i.e., the output of $\phi(\mathcal{D})$). The network computes:

\begin{equation}
    g_\theta(\mathbf{x}) = \mathbf{W}_2 \,\text{ReLU}(\mathbf{W}_1 \mathbf{x} + \mathbf{b}_1) + \mathbf{b}_2,
\label{eq:logits}
\end{equation}

where
\begin{itemize}
    \item $\mathbf{W}_1 \in \mathbb{R}^{64 \times d_\phi}$ and $\mathbf{b}_1 \in \mathbb{R}^{64}$ are the weight matrix and bias vector of the first layer,
    \item $\mathbf{W}_2 \in \mathbb{R}^{K \times 64}$ and $\mathbf{b}_2 \in \mathbb{R}^{K}$ are the weight matrix and bias vector of the second layer,
    \item $K$ is the number of label noise detectors in the ensemble,
    \item $d_\phi = d_{\text{in}}$ and $K = d_{\text{out}}$.
\end{itemize}

The output logits $g_\theta(\mathbf{x})$ are converted to a probability vector over detectors using softmax normalization:
\begin{equation}
    \mathbf{w} = \text{softmax}(g_\theta(\mathbf{x})),
\label{eq:weights}
\end{equation}
so that $\sum_{k=1}^K w_k = 1$ and $w_k$ represents the predicted importance of detector $k$ for the given dataset.

\subsubsection{Pairwise Ranking Loss}

The meta-model predicts detector logits $\hat{\mathbf{w}} = g_\theta(\mathbf{x})$, which are converted to weights via $\mathbf{w} = \text{softmax}(\hat{\mathbf{w}})$. The detector ROC-AUC values $\mathbf{a}^{(m)} \in \mathbb{R}^K$ for dataset $m$ serve as supervision targets, where $a_i^{(m)}$ is the performance of detector $i$.

We train the meta-model using a pairwise ranking loss, which penalizes disordered pairs of detectors:
\begin{equation}
\begin{split}
    \mathcal{L} = \frac{1}{K(K-1)} \sum_{i \neq j} \mathbb{E}_m \Big[ 
    &\max\Big(0, \; \text{margin} \\
    &- \text{sgn}(a_i^{(m)} - a_j^{(m)}) \cdot (\hat{w}_i^{(m)} - \hat{w}_j^{(m)}) \Big) \Big].
\end{split}
\label{eq:pairwise_loss}
\end{equation}

Intuitively, the loss minimizes the number and severity of rank inversions: it encourages detectors with higher ROC-AUC values to receive higher predicted logits than weaker detectors. This aligns the learned weights with relative detector quality without directly predicting raw ROC-AUC values.

\subsubsection{Training Protocol}
The meta-model is trained for 150 epochs with Adam ($\text{lr}=5{\times}10^{-3}$) using instances from OpenML-CC18~\cite{bischl2019openml}, with up to six per dataset (one per noise rate). Datasets with fewer than two classes are discarded. Features are standardized, and datasets are stratified and subsampled to at most 1,000 examples to preserve class balance. Meta-training pairs are constructed by simulating noisy datasets and benchmarking detector performance across data regimes (Algorithm~\ref{algorithm:meta_data}), and are then used to train a meta-model that predicts dataset-dependent detector weights for adaptive ensembling at inference (Algorithm~\ref{algorithm:meta_training}).

\begin{algorithm}[t]
\caption{Meta-Training Data Generation}\label{algorithm:meta_data}
\begin{algorithmic}[1]

\Require Benchmark dataset collection $\mathbf{\mathcal{D}} = \{\mathcal{D}_1, \dots, \mathcal{D}_J\}$, detector set $\mathcal{F}=\{f_1,\dots,f_K\}$, noise rates set $\mathcal{E}$, empty training pair list $\mathcal{T} \gets []$
\Ensure Meta-training pairs $\mathcal{T}$

\For{each dataset $\mathcal{D}_j \in \mathbf{\mathcal{D}}$}
    \State Compute dataset descriptor $\phi(\mathcal{D}_j)$
    \For{each noise rate $\varepsilon \in \mathcal{E}$}
        \State Generate noisy dataset $\tilde{\mathcal{D}}_j$ using Algorithm~\ref{alg:fringe}
        \For{each detector $f_k \in \mathcal{F}$}
            \State Train $f_k$ on $\tilde{\mathcal{D}}_j$
            \State Compute detector suspicion scores $\mathbf{s}_k$
            \State Evaluate ROC-AUC $a_k$ with the ground-truth noise mask
        \EndFor
        \State Append $(\phi(\tilde{\mathcal{D}}_j), \mathbf{a})$ to $\mathcal{T}$ 
    \EndFor
\EndFor

\State \Return $\mathcal{T}$ 

\end{algorithmic}
\end{algorithm}

\begin{algorithm}[h]
\caption{Meta-Model Training and Inference}
\label{algorithm:meta_training}
\begin{algorithmic}[1]

\Require Meta-training pairs $\mathcal{T}$, detector set $\mathcal{F}=\{f_1,\dots,f_K\}$, target dataset $\mathcal{D}$
\Ensure Final ensemble suspicion scores $\mathbf{s}^*$

\State \textbf{Meta-Model Training}
\State Initialize meta-model $g_\theta$
\For{each $(\mathbf{x}, \mathbf{a}) \in \mathcal{T}$}
    \State Compute predicted logits $\hat{\mathbf{w}} = g_\theta(\mathbf{x})$ (Eq.~\ref{eq:logits})
    \State Compute pairwise ranking loss with $\hat{\mathbf{w}}$ and $\mathbf{a}$ (Eq.~\ref{eq:pairwise_loss})
    \State Update $g_\theta$ via gradient step
\EndFor

\vspace{1mm}
\State \textbf{Inference on Target Dataset}
\State Extract descriptor $\mathbf{x} = \phi(\mathcal{D})$
\State Compute detector weights $\mathbf{w} = \text{softmax}(g_\theta(\mathbf{x}))$

\For{each detector $f_k \in \mathcal{F}$}
    \State Train detector $f_k$ on $\mathcal{D}$ and compute suspicion score $s_k$
\EndFor

\State Compute ensemble score $\mathbf{s}^* = \sum_{k=1}^{K} w_k s_k$
\State \Return $\mathbf{s}^*$

\end{algorithmic}
\end{algorithm}

\subsection{Baseline: Confident Learning}

As our primary baseline, we employ the Confident Learning framework implemented in the modern \texttt{cleanlab} library \cite{northcutt2021confident}. Confident Learning estimates the probability that an observed label is correct by leveraging cross-validated model predictions and the structure of the label noise distribution. Rather than explicitly implementing the original pruning-based pipeline, we follow the recommended scoring interface provided in Cleanlab v2, which produces continuous label-quality scores suitable for ranking-based evaluation.

First, out-of-sample predicted probabilities $\hat{P}(y\mid\mathbf{x})$ are estimated using stratified cross-validation with multinomial logistic regression. This produces a matrix of predicted probabilities for each example across all classes while avoiding training-set leakage. These probabilities are then passed to the Cleanlab ranking procedure, which computes per-example label quality scores based on the \textit{normalized margin} criterion.

The normalized margin measures the difference between the probability assigned to the observed label and the highest probability assigned to any alternative class. Intuitively, examples whose observed label receives substantially lower probability than a competing class are more likely to be mislabeled. Cleanlab converts this margin into a label quality score $q_i \in [0,1]$, where larger values correspond to more reliable labels. For compatibility with our evaluation protocol, we convert this quantity into a suspicion score $s_i = 1 - q_i$ so that higher values indicate a greater likelihood of mislabeling.


\subsection{Evaluation Protocol}

We evaluate performance on a diverse collection of benchmark datasets spanning standard tabular tasks and high-dimensional biomedical domains, following prior work on label noise detection~\cite{nazaretyan2025benchmarking}. The tabular datasets include Adult, DryBean, Magic, and Chess from OpenML~\cite{bischl2019openml}. To capture more challenging, high-dimensional regimes, we additionally include four medical datasets: two RNA expression datasets derived from single-cell transcriptomic data~\cite{pijuan2019single}, and two ClinVar-based variant classification datasets~\cite{landrum2018clinvar}.

The RNA datasets are constructed by selecting clusters that are distinguishable but not trivially separable, yielding moderately difficult classification tasks. The ClinVar datasets incorporate real-world label noise from evolving clinical interpretations of genetic variants, providing a realistic evaluation setting alongside synthetic noise~\cite{landrum2018clinvar}.

All datasets are subsampled to at most 1,500 instances to standardize computational cost. For controlled experiments, we inject fringe-based label noise at rate $\varepsilon$ and evaluate both the proposed meta-ensemble and the Confident Learning baseline.

\subsubsection{Metrics}

We report two primary metrics:

\begin{itemize}
    \item \textbf{Area Under the Receiver Operating Characteristic Curve (ROC-AUC)} evaluates how well the sample-level suspicion scores rank corrupted examples above clean examples. For the meta-ensemble, ROC-AUC is computed from the ensemble score \(s_i^*\) and the binary corruption indicator for each sample.
    \item \textbf{F1 score} is computed on a binary detection task by thresholding the top-$\lfloor N \cdot \varepsilon \rfloor$ samples, using the known noise budget. This assesses practical detection performance under a known corruption budget.
\end{itemize}

In addition to aggregate metrics, we log detector-specific ROC-AUC values to analyze the behavior of individual methods across datasets and noise regimes. This enables a detailed study of detector specialization and its relationship to dataset characteristics.

\subsection{Interpretability via SHAP}
To quantify which dataset properties most strongly drive detector preference, we compute Shapley attributions over the softmax-normalized detector weight outputs. Specifically, we apply Kernel SHAP~\cite{lundberg2017unified} to the trained meta-model. A background set of 20 randomly sampled training descriptors is used as the reference distribution. SHAP values are computed over up to 40 held-out descriptors, and mean absolute SHAP values are averaged across all $K$ detector outputs to produce a single global importance score per meta-feature.

\section{Experiments and Results}
\subsection{Overall Performance Metrics}
\begin{figure}[t]
    \centering
    \includegraphics[width=1\linewidth]{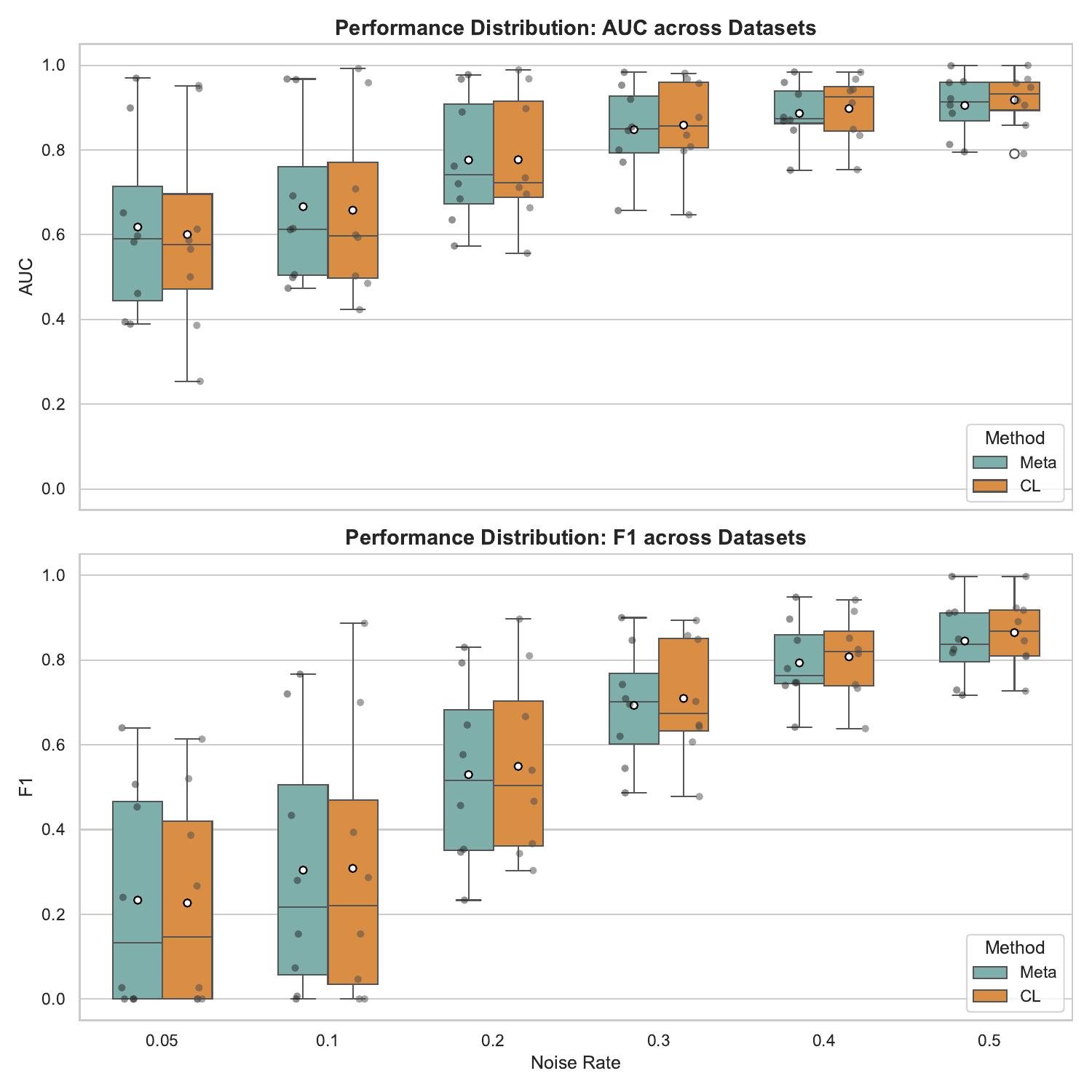}
    \caption{ROC-AUC and F1 Comparison of the Meta-Ensemble and Confident Learning.}
    \label{fig:1}
\end{figure}
 
Across all datasets and noise levels, both the proposed meta-ensemble and the Confident Learning baseline achieve moderate-to-high ROC-AUC under the tested conditions, with performance generally increasing as the noise rate grows. This trend, as shown in Figure~\ref{fig:1}, is expected because higher noise levels produce more mislabeled examples that are easier for detectors to identify. The distributional boxplots show that the meta-ensemble achieves performance comparable to or slightly exceeding Confident Learning across most regimes. In particular, the meta-ensemble shows modest improvements in median ROC-AUC and F1 at moderate noise levels (e.g., 0.20–0.40), where label ambiguity is most pronounced. At very low noise rates, both methods demonstrate higher variance due to the limited number of corrupted labels available for detection. Overall, these results indicate that the adaptive ensemble is competitive with a strong single-method baseline while maintaining stable performance across diverse datasets and noise regimes.

\begin{figure}[t]
    \centering
    \includegraphics[width=\linewidth]{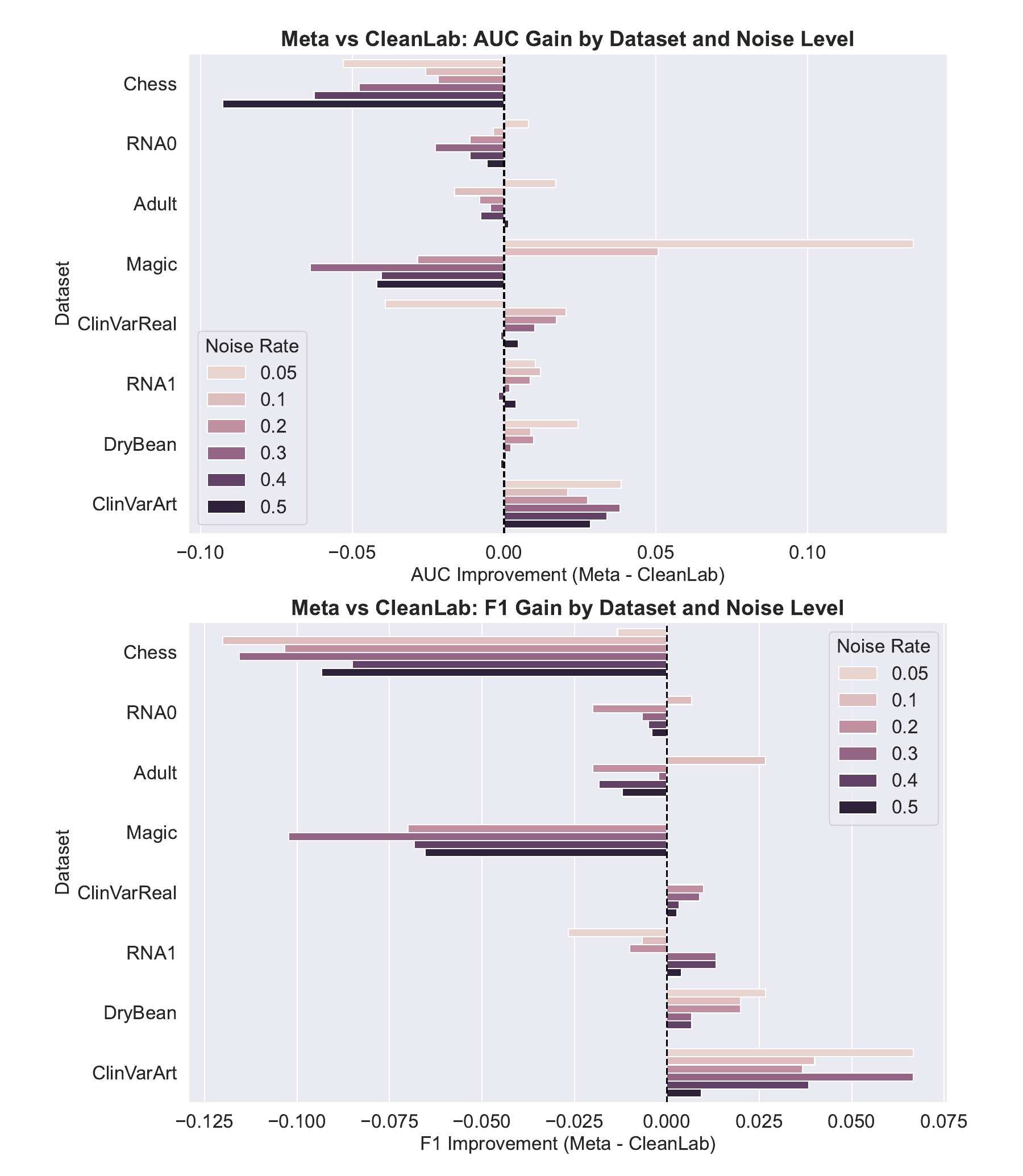}
    \caption{Meta-Ensemble and Confident Learning across Datasets.}
    \label{fig:2}
\end{figure}

To understand when the meta-ensemble provides benefits over the baseline, we analyze performance differences at the level of individual datasets and noise rates, rather than relying solely on aggregate metrics. This allows us to identify regimes where one method consistently outperforms the other. Figure~\ref{fig:2} reports dataset-level and noise-rate-specific performance differences between the proposed method and Confident Learning. Positive values indicate that the meta-ensemble achieves higher detection accuracy. The results reveal substantial heterogeneity in relative performance across both datasets and noise regimes.

ClinVarArt exhibits the strongest and most consistent improvements under the meta-ensemble, with gains increasing at higher noise rates, particularly for F1 score, where improvements exceed 0.05 at the 0.4 and 0.5 noise levels. DryBean and RNA1 similarly show moderate positive gains across most noise conditions. In contrast, Chess demonstrates a consistent advantage for Confident Learning across all noise rates and both metrics, with the gap most pronounced at lower noise levels. Magic and RNA0 also favor the baseline, though less uniformly.

Interestingly, several datasets exhibit varied noise-rate-dependent behavior. For example, Adult shows strong meta-ensemble improvements at specific noise rates (e.g., 0.1 for ROC-AUC, 0.2 for F1) but performs worse than the baseline at other levels. This indicates that the relative effectiveness of adaptive ensembling versus single-model confidence scoring depends not only on dataset structure but also on the severity and concentration of label corruption.

\subsection{Performance of Individual Detectors across Datasets}
\begin{figure}[t]
    \centering
    \includegraphics[width=1\linewidth]{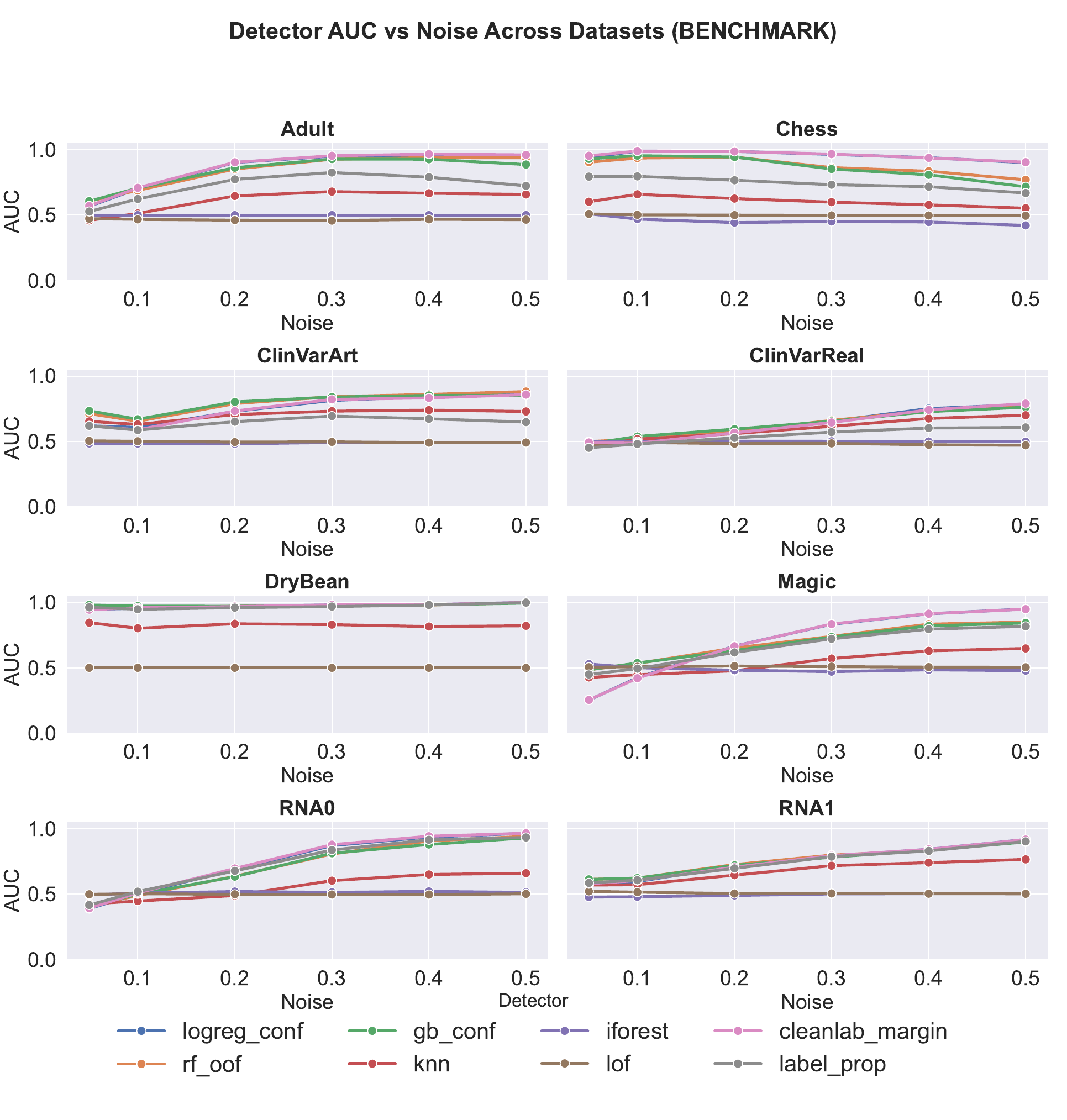}
    \caption{Performance of Meta-Ensemble Component Detectors across Noise Levels.}
    \label{fig:3}
\end{figure}

Figure~\ref{fig:3} shows the behavior of individual detectors across benchmark datasets and noise regimes to understand the sources of ensemble performance. Confidence-based detectors from supervised classifiers—logistic regression, gradient boosting, and random forests—generally achieve high ROC-AUC values, particularly on datasets where mislabeled examples induce noticeable probabilistic disagreement. However, their performance varies across datasets and noise levels, underscoring that no single detector uniformly dominates. In contrast, unsupervised anomaly detectors, such as Isolation Forest and Local Outlier Factor, show limited overall effectiveness on this specific benchmark, though they can detect mislabeled samples in specific high-variance or low-density regions. The K-nearest-neighbor disagreement detector provides moderate detection performance, contributing in certain datasets (ClinVarReal, ClinVarArt, RNA0, and RNA1), where local label inconsistencies are prominent. These results highlight substantial heterogeneity in detector effectiveness, motivating the proposed meta-learning framework that adaptively combines detector outputs rather than relying on a single fixed strategy.

\subsection{Meta-Model Descriptor Importance}
\begin{figure}[t]
    \centering
    \includegraphics[width=1\linewidth]{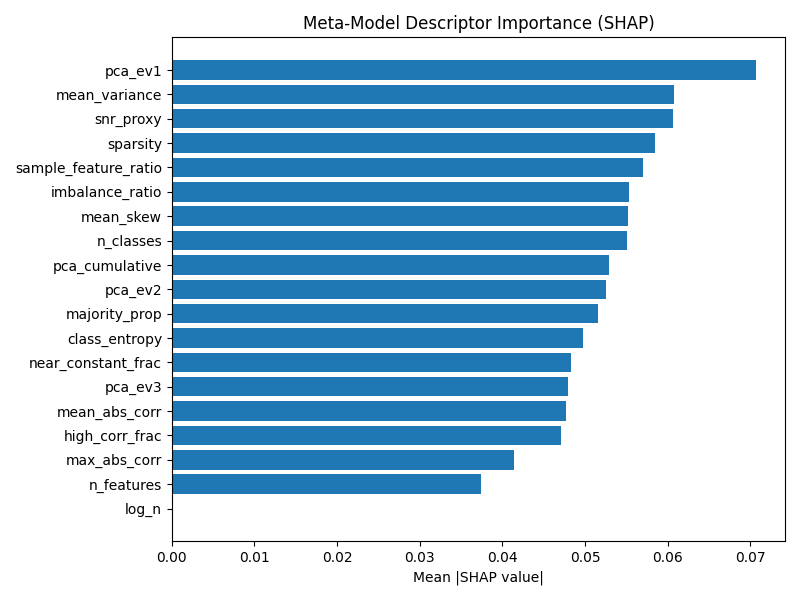}
    \caption{Analysis of SHAP Values for Dataset Descriptors.}
    \label{fig:4}
\end{figure}

We use SHAP values \cite{lundberg2017unified} to assess how dataset descriptors influence the meta-model's predicted detector weights. Figure~\ref{fig:4} shows that descriptors capturing signal structure and feature-space characteristics dominate, rather than raw dimensionality. The first principal component's explained variance (\texttt{pca\_ev1}) has the highest attribution (\~0.07), followed by mean feature variance and a signal-to-noise proxy, indicating that detector selection favors datasets with concentrated informative variance. Sparsity ranks fourth, suggesting that feature density may also matter.

Descriptors of intermediate importance include the sample-to-feature ratio, number of classes, imbalance ratio, and mean feature skewness. Correlation structure descriptors and raw dimensionality measures, including \texttt{log\_n}, have minimal impact.  

Overall, SHAP values range roughly 0.045–0.07, suggesting the meta-model integrates multiple characteristics rather than relying on one. PCA-based features highlight that adaptive weighting is primarily guided by intrinsic variance structure, with simpler scale features playing a secondary role.

\subsection{Dataset Descriptor Heatmap}

\begin{figure}[t]
    \centering
    \includegraphics[width=1\linewidth]{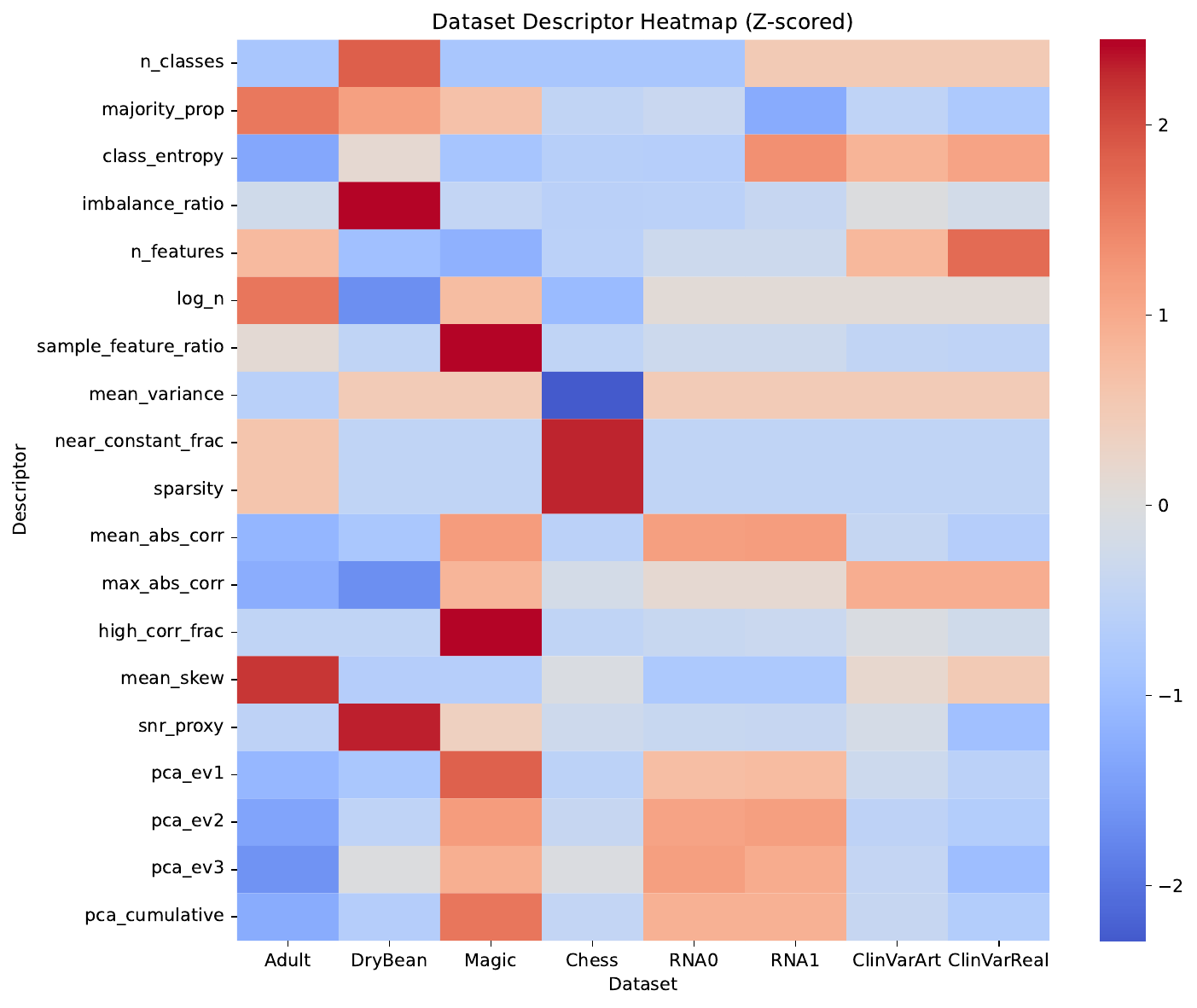}
    \caption{Dataset Descriptor Deviations from the Mean (z Scores).}
    \label{fig:5}
\end{figure}

To understand how dataset characteristics may influence detector performance, we analyze the distribution of descriptor values across benchmark datasets. We compute z-scored profiles for each descriptor to standardize scales and highlight relative differences. Figure~\ref{fig:5} shows the resulting heatmap, revealing heterogeneity in dataset structure: Chess exhibits high sparsity and many near-constant features; Magic has a high sample-to-feature ratio; DryBean shows extreme class imbalance; and the RNA datasets display high dimensionality with few samples. This diversity helps explain why different detection strategies perform variably across datasets and motivates the use of an adaptive meta-ensemble.

\subsection{Descriptor Correlations with Overall Model Performance}

\begin{figure}[t]
    \centering
    \includegraphics[width=1\linewidth]{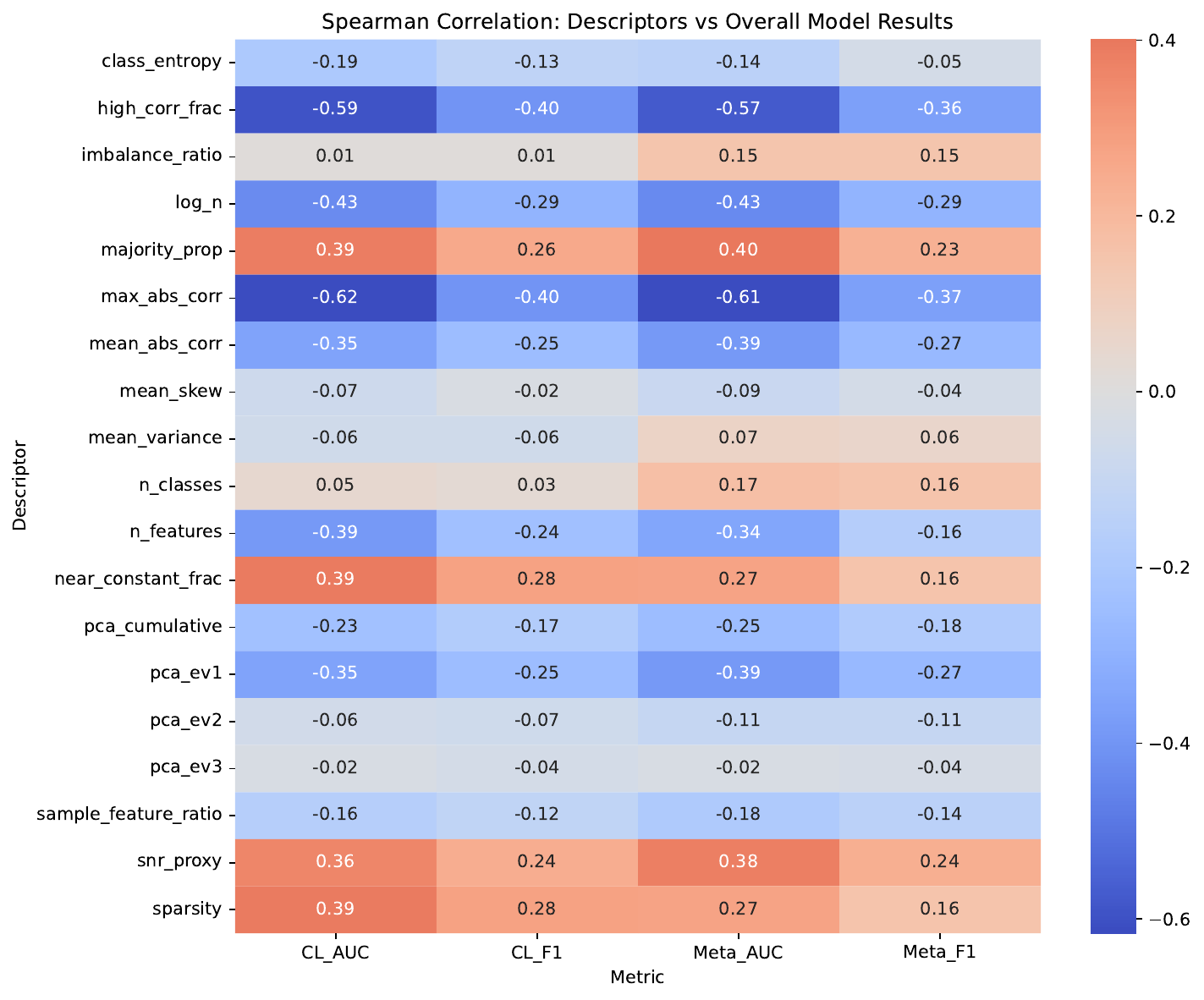}
    \caption{Dataset Descriptor Correlations with Overall Performance.}
    \label{fig:6}
\end{figure}

To understand which dataset characteristics influence label noise detection, we examine correlations between descriptor values and overall model performance. 

Figure~\ref{fig:6} shows these relationships for both Confident Learning and the meta-ensemble. Detection performance tends to decrease (negative correlation) in high-dimensional datasets with strong feature correlations (\texttt{high\_corr\_frac}, \texttt{max\_abs\_corr}, \texttt{pca\_ev1}, \texttt{log\_n}, \texttt{n\_features}), and improves (positive correlation) in sparse datasets with many near-constant features, dominant majority classes, and high signal-to-noise ratios. Interestingly, these patterns are consistent across both methods, suggesting that despite architectural differences, they are challenged by similar dataset regimes. This analysis highlights which dataset properties are most strongly associated with  difficulty in label-noise detection.

\subsection{Descriptor Correlations with Detector Performance}

\begin{figure}[t]
    \centering
    \includegraphics[width=1\linewidth]{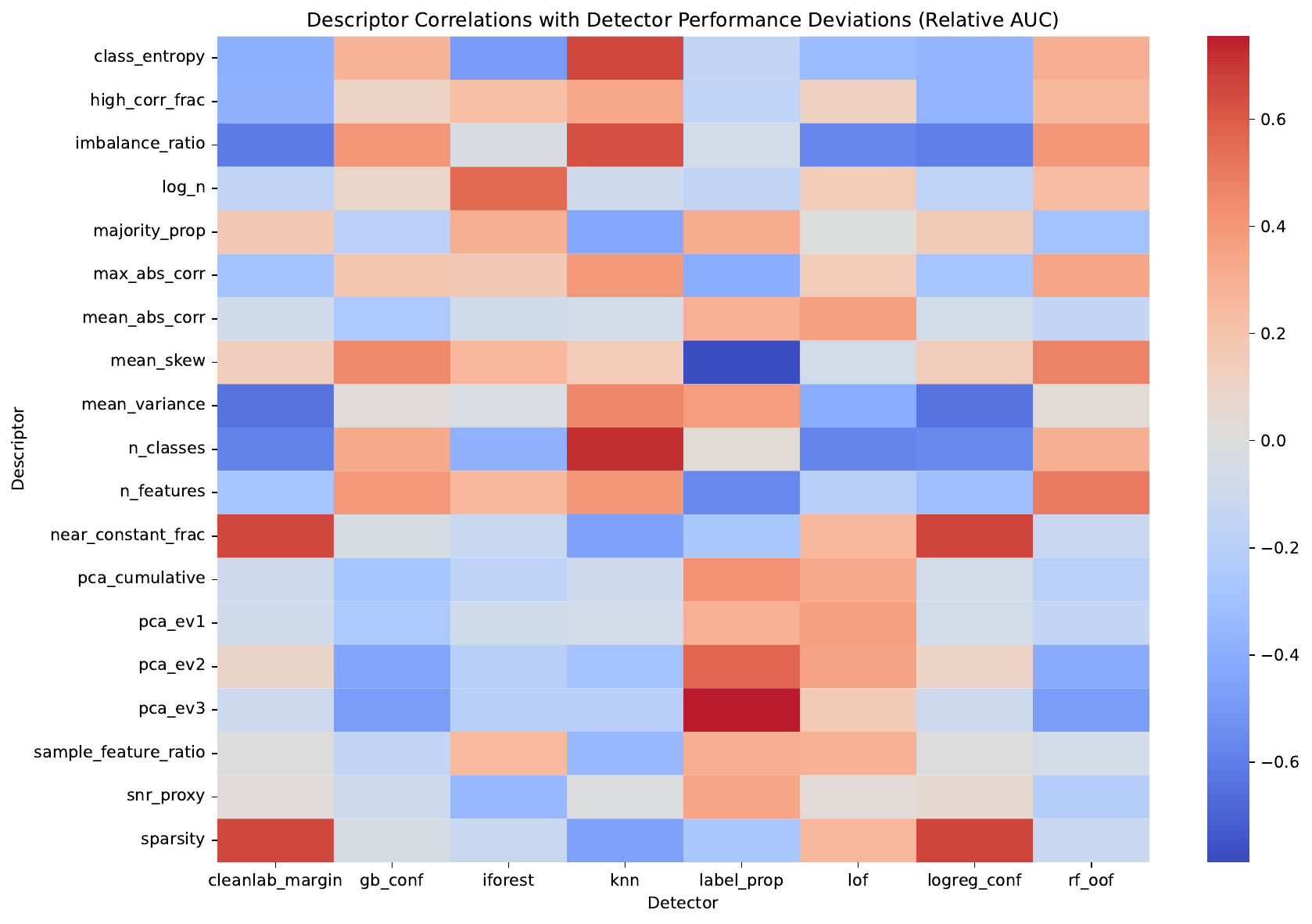}
    \caption{Descriptor Correlations with Detector-Specific Performance.}
    \label{fig:7}
\end{figure}

To investigate how individual dataset characteristics influence specific detectors, we examine correlations between dataset descriptors and detector-level performance. 

Figure~\ref{fig:7} visualizes these relationships, revealing strong specialization patterns. Sparsity and near-constant features favor confidence-based methods (\texttt{cleanlab\_margin}, \texttt{logreg\_conf}) while reducing effectiveness for tree-based and nearest-neighbor detectors. Class imbalance (\texttt{imbalance\_ratio}, \texttt{class\_entropy}) benefits KNN disagreement, likely because local neighborhoods become more homogeneous in imbalanced settings. Feature skewness (\texttt{mean\_skew}) has the most pronounced detector-specific impact, sharply degrading label propagation performance while leaving other methods largely unaffected, likely due to its reliance on smooth, well-behaved feature geometries. These divergent responses highlight that no single detector dominates across all dataset regimes, motivating the adaptive ensemble framework: the meta-model can learn to upweight detectors whose inductive biases align with observed dataset characteristics, exploiting complementary signals inaccessible to any fixed strategy.

\section{Discussion}
The experimental results provide strong evidence that detector effectiveness is highly dependent on dataset structure and that adaptive ensembling offers an effective way to exploit this heterogeneity. Although the meta-ensemble and Confident Learning achieve largely indistinguishable overall performance, the detailed descriptor and detector analyses reveal substantial differences in how each method leverages dataset characteristics.

Specifically, the dataset-descriptor heatmap (Figure \ref{fig:5}) and overall correlation analysis (Figure \ref{fig:6}) demonstrate that both the meta-ensemble and Confident Learning are sensitive to similar global dataset properties. In particular, high dimensionality, strong feature correlations, and dominant principal components (high \texttt{pca\_ev1}) are consistently associated with reduced detection performance. These regimes likely correspond to datasets where decision boundaries are more complex or where feature redundancy obscures clear class separation. Conversely, sparse datasets with near-constant features and higher signal-to-noise ratios are more favorable for label error detection, as mislabeled points stand out more clearly against the underlying structure.

However, the individual detector-level analysis (Figures \ref{fig:3} and \ref{fig:7}) reveals a more nuanced picture. Different detectors respond in fundamentally different ways to the same dataset characteristics. Confidence-based methods (e.g., logistic regression and Confident Learning) benefit strongly from sparsity and low-variance features, where probabilistic models can form stable decision boundaries. In contrast, KNN-based methods show improved relative performance under class imbalance, likely due to increased local homogeneity in majority-class neighborhoods. Label propagation is highly sensitive to feature skewness, with performance degrading sharply as skew increases, indicating its reliance on well-behaved, near-symmetric feature geometries. Meanwhile, unsupervised anomaly detectors (Isolation Forest, Local Outlier Factor) remain largely uninformative across most data regimes, reinforcing the observation that label noise does not necessarily manifest as feature-space outliers.

These findings support the core premise under the evaluated conditions: no single detector is universally optimal, and their relative performance varies systematically with the observed dataset meta-properties. The proposed meta-model leverages this structure by learning a mapping from dataset descriptors to detector weights, effectively performing a form of data-driven algorithm selection. Rather than treating detector outputs as equally informative, the ensemble adapts to emphasize detectors whose inductive biases align with the observed dataset characteristics.

Importantly, the similarity in global performance correlations (Figure \ref{fig:6}) between the meta-ensemble and Confident Learning suggests that both methods are constrained by the same fundamental limits of the data. The advantage of the meta-ensemble, therefore, is not in overcoming these limits entirely, but in better navigating intermediate regimes where multiple weak signals can be combined. This is most evident in datasets, such as ClinVarArt and DryBean, where heterogeneous noise patterns allow the ensemble to outperform any single detector. Conversely, Chess and Magic represent regimes where Confident Learning's single-model probabilistic approach is already near-optimal. Chess exhibits extreme sparsity and a high fraction of near-constant features (Figure \ref{fig:5}), conditions under which confidence-based detectors already achieve maximal discriminability and additional ensemble members contribute primarily noise rather than complementary signal. Similarly, Magic's high sample-to-feature ratio and strong class separation create a regime where logistic regression confidence alone provides sufficient information for label error detection. In these cases, the meta-ensemble's adaptive weighting cannot improve upon an already-strong baseline, and the inclusion of unsuitable detectors dilutes the final score. This highlights a fundamental limitation of ensemble approaches: when a single strong signal dominates, aggregation offers no benefit and may introduce unnecessary variance.

Overall, these results underscore the value of a data-centric perspective on AI-assisted data science in label noise detection. By explicitly capturing the relationship between dataset characteristics and detector effectiveness, our adaptive ensemble provides a flexible, interpretable, and automated framework for improving robustness in noisy supervision scenarios, supporting more reliable and scalable data science workflows.
\section{Conclusion}

In this work, we introduced an adaptive ensemble framework for label noise detection in tabular datasets, formulated as a meta-learning problem over dataset characteristics. By combining a diverse set of complementary detectors and learning to weight them based on dataset-level descriptors, the approach addresses a key challenge in data-centric AI: no single detection method performs optimally across all data regimes.

Empirical evaluation across benchmark datasets and controlled noise levels shows that the meta-ensemble is competitive with and, in some cases, outperforms the baseline. 
More importantly, our analysis demonstrates that detector effectiveness is associated with dataset properties, such as sparsity, class imbalance, feature correlation, and signal structure. The meta-model successfully captures these relationships, enabling adaptive weighting that aligns detector inductive biases with dataset characteristics, a capability particularly relevant for AI systems that must operate across heterogeneous datasets.
Beyond performance gains, this work provides a data-centric and interpretable perspective on label noise detection. Through descriptor analysis, SHAP-based interpretability, and detector-level correlation studies, we offer insight into why certain methods succeed or fail under different conditions, reframing label quality assessment as a function of dataset structure rather than purely model choice. This aligns closely with the goals of AI data science systems, which aim to automate, assist, and optimize data workflows while providing transparency to human practitioners.

We note several limitations. The framework introduces additional computational overhead, as multiple detectors and descriptor extraction must be evaluated, which may constrain deployment in large-scale or low-latency settings. However, this overhead can be reframed as a cost-aware trade-off, where the meta-model selects only the top-ranked detector instead of a full ensemble, improving efficiency while preserving most performance. The meta-model is trained on a relatively small collection of benchmark datasets, which may limit generalization to domains with substantially different feature distributions or noise patterns. However, the descriptor-driven approach suggests a modality-agnostic extension, where appropriate dataset descriptors could enable transfer beyond tabular settings, such as in image- or text-based data. Additionally, our evaluation relies on synthetic fringe-based noise injection, which, while more realistic than uniform corruption, may not fully capture the complexity of real-world labeling across all contexts. Nevertheless, the strong performance observed on the ClinVar and ClinVarArt datasets~\cite{landrum2018clinvar}, both of which contain real-world and clinically derived annotation noise, suggests that medical labeling errors may still exhibit fringe-like structure, where uncertainty concentrates near decision boundaries due to annotator disagreement.


Overall, this study demonstrates that adaptive, descriptor-driven ensembling is a promising direction for robust label noise detection, bridging the gap between algorithm-centric and data-centric approaches in modern machine learning. By emphasizing dataset-aware reasoning, this work contributes to the development of AI data-science systems capable of supporting end-to-end data workflows in real-world settings.


\balance
\bibliographystyle{ACM-Reference-Format}
\bibliography{references}

\end{document}